\documentclass[10pt, a4paper]{article}
\usepackage{lrec2000}
\usepackage{graphicx}

\title{Test Collections for Patent-to-Patent Retrieval and \\
  Patent Map Generation in NTCIR-4 Workshop}

\name{Atsushi Fujii$^{\ast}$, Makoto Iwayama$^{\dagger}$, Noriko Kando$^{\ddagger}$}

\address{$^{\ast}$Institute of Library and Information Science \\
University of Tsukuba \\
1-2 Kasuga, Tsukuba, 305-8550, Japan \\
fujii@slis.tsukuba.ac.jp \\ \\
$^{\dagger}$Hitachi, Ltd. \\
1-280 Higashi-Kougakubo, Kokubunji, 185-8601, Japan \\
iwayama@crl.hitachi.co.jp \\ \\
$^{\ddagger}$
National Institute of Informatics \\
2-1-2 Hitotsubashi, Chiyoda-ku, 101-8430, Japan \\
kando@nii.ac.jp}

\abstract{This paper describes the Patent Retrieval Task in the Fourth
  NTCIR Workshop, and the test collections produced in this task. We
  perform the invalidity search task, in which each participant group
  searches a patent collection for the patents that can invalidate the
  demand in an existing claim. We also perform the automatic patent
  map generation task, in which the patents associated with a specific
  topic are organized in a multi-dimensional matrix.}

\begin{document}

\maketitleabstract

\section{Introduction}
\label{sec:introduction}

In the Third NTCIR Workshop (NTCIR-3), which is a TREC-style
evaluation forum for research and development on information retrieval
and natural language processing, the authors of this paper organized
the Patent Retrieval
Task~\cite{iwayama:sigir-2003,iwayama:ntcir-2003}. This was the first
serious effort to produce a test collection for evaluating patent
retrieval systems.

The process of patent retrieval differs significantly depending on the
purpose of retrieval. In NTCIR-3 Workshop, the ``technology survey''
task was performed, in which patents are regarded as technical
publications rather than legal documents.
In practice, given a query, which is a clipping of a newspaper
articles related to a specific technology, two years of patent
publications were searched for the documents relevant to the query.
Search topics were in five languages. The same contents in Japanese,
English, Korean, traditional/simplified Chinese were used to perform
cross-language retrieval.

Given a success in NTCIR-3 Workshop, the authors are also performing
the Patent Retrieval Task in NTCIR-4 Workshop, which is held from
January 2003 to June 2004. However, unlike NTCIR-3 Workshop, we are
focusing on the ``invalidity search'' and ``patent map generation''
tasks.  This paper describes the test collections for both tasks.

Because NTCIR-4 Workshop is performed in one and half years, it is
difficult to explore long-term research topics, such as the patent map
generation task.  Thus, while we perform the invalidity search task,
which resembles the traditional ad-hoc IR task, as the main task, we
perform the patent map generation task as a feasibility study, for
which no quantitative evaluation is conducted.

\section{Invalidity Search Task}

\subsection{Overview}

The purpose of invalidity search is to find the patents that can
invalidate the demand in an existing claim. This is an associative
patent (patent-to-patent) retrieval task. In real world, invalidity
search is usually performed by examiners in a government patent office
and searchers of the intellectual property division in private
companies.

The task was performed as follows. First, the task organizers (i.e.,
the authors of this paper) provided each participant group with the
document collection and search topics.

Second, each group submitted retrieval the results queried by the
topics.  In a single retrieval result, the top 1000 retrieved
documents must be sorted by the relevance score. However, because
patent documents are long, it is effective to indicate the important
passages (i.e., fragments) in a relevant document. Thus, for each
retrieved document, all passages in the document must be sorted as to
which a passage provides grounds to judge if the document is relevant.

Third, human experts performed relevance judgment for the submitted
results and produced a list of relevant documents and passages, on a
topic-by-topic basis. Finally, the list was used to evaluate each
submitted result.

In the dry run, which was performed from June to September in 2003,
seven topics were produced and used for a preliminary evaluation.  In
the formal run, 103 search topics were produced and the evaluation
results for each group will be released at the workshop final meeting
in June 2004.  The analysis of the formal run results has not been
completed and is beyond the scope of this paper.

After the workshop final meeting, we complete a test collection
consisting of the search topics, the document collection, and the
relevance judgments for each topic.

\subsection{Document Sets}

The document set used as a target collection consists of five years of
unexamined Japanese patent applications published in 1993-1997. The
file size and number of documents are approximately 40GB and 1.7M,
respectively.

For the sake of passage-based evaluation, the passages in each
document were standardized. In Japanese patent applications,
paragraphs are identified and annotated with the specific tags by
applicants. We used these paragraphs as passages, and therefore the
passage identification process was fully automated.

The English patent abstracts, which are human translations of the
Japanese Patent Abstracts published in 1993-1997, were also provided
for training English-to-Japanese cross-language IR systems.

\subsection{Search Topics}

A search topic is a Japanese patent application rejected by the
Japanese Patent Office.  For each topic patent, one or more citations
were identified by examiners to invalidate the demand in the topic
patent. If these citations are included in our document collection,
they can be used as relevant documents for the topic.

We asked 12 members of the Intellectual Property Information Search
Committee in the Japan Intellectual Property Association (JIPA) to
produce seven topics for the dry run and 34 topics for the formal run.
Each JIPA member belongs to the intellectual property division in the
company he or she works for, and they are all experts in patent
searching. The JIPA member also performed relevance judgment to
enhance the relevant documents.

A search topic file includes a number of additional SGML-style tags.
The claim as a target of invalidation is specified by \verb|<CLAIM>|.

A claim usually consists of multiple components (e.g., parts of a
machine and substances of a chemical compound) and relevance judgment
is performed on a component-by-component basis in real world case. To
simulate this scenario, human experts annotate each component with
\verb|<COMP>|.

To invalidate an invention in a topic patent, relevant documents must
be the ``prior art'', which had been open to the public before the
topic patent was filed. Thus, the date of filing is specified by
\verb|<FDATE>| and only the patents published before the topic was
filed can potentially be relevant.

To perform cross-language retrieval, the claims translated into
English and simplified Chinese are also used. Thus, the topic language
is specified by \verb|<LANG>|.  However, the translated claims do not
maintain the order of phrases and sentences in Japanese claims and
thus do not include the \verb|<COMP>| tags.  Figure~\ref{fig:topic}
shows an example topic claim translated into English.

\begin{figure}[htbp]
  \begin{center}
    \leavevmode
    \small
    \begin{quote}
      \verb|<TOPIC>|\\
      \verb|<NUM>008</NUM>| \\
      \verb|<CLAIM>|(Claim 1) A sensor device, characterized in that
      an open recessed part is formed on a box-shaped forming base, a
      conductive film of a designated pattern is formed on the surface
      of the forming base including the inner surface of the recessed
      part, an element for a sensor is bonded to the recessed part,
      and the forming base is closed with a cover.\verb|</CLAIM>| \\
      \verb|</TOPIC>|
    \end{quote}
    \caption{The claim in an English search topic (008).}
    \label{fig:topic}
  \end{center}
\end{figure}

Through a preliminary study in collaboration with JIPA, we found that
for invalidity search the number of relevant documents for a single
topic is small, compared with existing IR test collections.
Consequently, the evaluation results obtained with our collection can
potentially be unstable.

The same problem is identified in the question answering task, and
thus the hundreds of questions are usually used to resolve this
problem~\cite{voorhees:sigir-2000}.

To increase the number of topics with a limited cost, we produced
additional 69 topics for which only the citations provided by the
Japanese Patent Office were used as the relevant documents. However,
the validity of rejection was verified manually,  the process of
producing additional topics was not fully automated.

\subsection{Submissions}

Each group was allowed to submit one or more retrieval results, in
which at least one result must be obtained using only the
\verb|<CLAIM>| and \verb|<FDATE>| fields. For the remaining results,
any information in a topic file, such as the International Patent
Classification (IPC) codes, can be used.

The results of the dry run showed that for specific topics, an
IPC-base system successfully retrieved relevant patents that could not
be retrieved by the text-based systems.

\subsection{Relevance Judgments}

The relevance degree of a document with respect to a topic is
determined on the basis of the relevance degrees of the document with
respect to each component in the topic. Relevance judgment for patents
is performed based on the following two ranks:
\begin{itemize}
\item patent that can invalidate a topic claim (A)
\item patent that can invalidate a topic claim, when used with other
  patents (B)
\end{itemize}
The documents that can invalidate the demands of all essential
components in a target claim were judged as ``A''. The documents that
can invalidate demands of most of the essential components in a target
claim (but not all essential components) were judged as ``B''.

For the main 34 topics, to identify relevant documents exhaustively,
the pooling method and manual search were used. The human experts who
produced the topics performed manual searches to collect as many
relevant patents as possible. The experts were allowed to use any
systems and resources, so that we were able to obtain a patent
document set retrieved under the circumstances of their daily patent
searching. The citations provided by the Japanese Patent Office were
also used as the relevant documents.

\begin{figure}[htbp]
  \bigskip
  \begin{center}
    \leavevmode
    \includegraphics[height=1.5in]{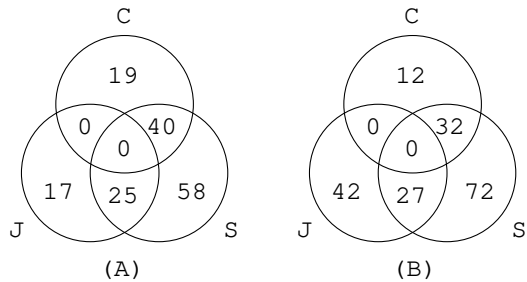}
    \caption{Details of the number of relevant documents.}
    \label{fig:diagram}
  \end{center}
\end{figure}

For the 34 topics, the resultant number of A and B documents were 159
and 185, respectively.  We analyzed details of the number of relevant
documents obtained by the different sources. In
Figure~\ref{fig:diagram}, ``C'', ``J'', and ``S'' denote the sets the
relevant documents (A and B) obtained by the citations, the manual
searches by the JIPA members, and the 30 systems participated in the
pooling, respectively.

It should be noted that because the JIPA members collected the
citations before the manual search, $|$C $\cap$ J$|$ is always zero.
Looking at this figure, each source was independently effective to
collect the relevant documents. While $|$C$|$ and $|$J$|$ were almost
equivalent, $|$S$|$ was comparable with \mbox{$|$C $\cup$ J$|$}.

The evaluation score is fundamentally determined by the conventional
mean average precision. At the same time, each group is encouraged to
propose new evaluation measures effective for patent IR systems.

In addition to the conventional document-based evaluation, we also
explore the passage-based evaluation. Relevant passages were
determined based on the following criteria:
\begin{itemize}
\item If a single passage can be grounds to judge the document in
  question as relevant (either A or B), this passage is judged as
  relevant.
\item If a ``group'' of passages can be grounds to judge the document
  in question as relevant, this passage group is judged as relevant.
\end{itemize}
The experts exhaustively identified all relevant passages and passage
groups.

It should be noted that a relevant passage group is equally
informative as a single relevant passage. In other words, we newly
introduce the concept of ``combinational relevance''.

This feature provides a salient contrast to the conventional IR
evaluation method, in which all relevant passages or documents are
independently important and thus combinations of partially relevant
documents are not considered.

The evaluation score for each system is determined by the number of
passages which would have to be searched until a user obtains a
sufficient grounds to judge the document as relevant.

\section{Patent Map Generation Task}

In principle, the purpose of the patent map generation task is to
generate a patent map driven by a specific theme, such as automobiles,
by (semi-)automatic method. This can be seen as a text mining task.

In practice, the organizers provided participants with the patent
documents retrieved by a specific topic, and participants are
requested to organize those documents in a two-dimensional matrix.
The x and y axes can vary depending on the topic, but they are usually
``problems to be solved'' and ``solutions'', respectively.

To produce the topics and documents, we used the test collection
produced for the NTCIR-3 Patent Retrieval Task. We selected six search
topics for which more than 100 relevant documents were identified.
The NTCIR-3 collection includes the following three document sets:
\begin{itemize}
\item two years worth of unexamined Japanese patent applications
  published in 1998 and 1999,
\item Japanese abstracts, the JAPIO Patent Abstracts, which are
  human-edited abstracts for the above applications,
\item English abstracts, the Patent Abstracts of Japan (PAJ), which
  are human translations of the JAPIO Patent Abstracts.
\end{itemize}
Any document set can be used for patent map generation purposes.
Because the search topics are in the five languages independently (see
Section~\ref{sec:introduction}), cross-language patent map generation
can also be performed.

However, the patent map generation task is as a feasibility study, and
thus human experts evaluated the submitted maps subjectively.

\section{Conclusion}

We built test collections for the patent-to-patent invalidity search
and automatic patent generation tasks in the NTCIR-4 Workshop.  After
the NTCIR-4 final meeting, the test collection will be available to
the public for research
purposes\footnote{http://www.slis.tsukuba.ac.jp/\~{}fujii/ntcir4/cfp-en.html}.

The test collections can directly be used for the following research
purposes:
\begin{itemize}
\item retrieval of very long semi-structured documents,
\item associative document retrieval,
\item passage retrieval,
\item evaluation of retrieval systems on the basis of combinational
  relevance,
\item classification and text mining.
\end{itemize}

Future work would include exploiting patent documents in different
applications, as follows:
\begin{itemize}
\item term recognition
  
  patent documents are associated with inventions and thus include a
  large number of new and technical terms.

\item sub-language studies
  
  claims in patent applications are written in a controlled language.

\item machine translation and cross-language retrieval
  
  inventions filed in multiple languages (i.e., patent families) can
  be used to extract translation lexicons.

\end{itemize}

\section{Acknowledgments}

The authors would like to thank the Japan Intellectual Property
Association for their support in the NTCIR-4 Patent Retrieval Task.

\bibliographystyle{lrec2000}

\end{document}